\documentclass[11pt,a4paper]{article}
\usepackage[hyperref]{eacl2021}
\usepackage{times}
\usepackage{latexsym}

\usepackage{microtype}
\usepackage{url}
\usepackage{CJKutf8}
\usepackage{multirow}
\usepackage{graphicx}
\usepackage{caption}
\usepackage{subcaption}
\usepackage{color}

\aclfinalcopy 

\title{Bilingual Dictionary-based Language Model Pretraining for Neural Machine Translation}

\author{
  Yusen Lin* \\
  Univ. of Maryland \\
  \scalebox{0.8}{\texttt{yusenlin@umd.edu}} \\ \And
  Jiayong Lin \\
  Univ. of Maryland \\
  \scalebox{0.8}{\texttt{jiayong@umd.edu}} \\ \And
  Shuaicheng Zhang \\
  Univ. of Maryland \\
  \scalebox{0.8}{\texttt{zshuai8@umd.edu}} \\ \And
  Haoying Dai \\
  Univ. of Maryland \\
  \scalebox{0.8}{\texttt{dhy@terpmail.umd.edu}} \\
}

\date{}

\begin{document}
\begin{CJK*}{UTF8}{gbsn}
\maketitle
\begin{abstract}
Recent studies have demonstrated a perceivable improvement on the performance of neural machine translation by applying cross-lingual language model pretraining \cite{Lample2019CrosslingualLM}, especially the Translation Language Modeling (TLM). 
To alleviate the need for expensive
parallel corpora by TLM, in this work, we incorporate the translation information from dictionaries into the pretraining process and propose a novel Bilingual Dictionary-based Language Model (BDLM).
We evaluate our BDLM in Chinese, English, and Romanian.
For Chinese-English, we obtained a 55.0 BLEU on WMT-News'19 \cite{tiedemann2012parallel} and a 24.3 BLEU on WMT'20 news-commentary, outperforming the Vanilla Transformer \cite{vaswani2017attention} by more than 8.4 BLEU and 2.3 BLEU, respectively. 
According to our results, the BDLM also has advantages on convergence speed and predicting rare words. 
The increase in BLEU for WMT'16 Romanian-English also shows its effectiveness in low-resources language translation. 
\end{abstract}

\section{Introduction}
Many previous works in language representation models like BERT \cite{devlin2018bert} have shown that prior knowledge is of great importance to many downstream tasks such as natural language understanding (NLU) and neural machine translation (NMT).
As one of the most successful Language Models (LMs), cross-lingual language models (XLMs) \cite{Lample2019CrosslingualLM}, 
which construct LM upon corpora from multiple languages, have demonstrated their ability to reach state-of-the-art performance in NLU as well as NMT. 
However, XLMs like Translation Language Model (TLM) require expensive corpus with parallel sentences for training and such a dataset might not be available in certain languages.

Meanwhile, a dictionary  contains information like Part-of-Speech (POS) and synonyms that human frequently refers to.
Inspired by this, there have been a few previous works, such as \citet{Zhang2016} and \citet{adams2017cross}, that incorporate bilingual dictionaries into NMT to improve translation quality and to obtain translations of rare words. 
\citet{Zhang2016} integrates the dictionary information into the training process by mixing words or characters to the corresponding tokens or synthesizing parallel sentences.
This method inevitably introduces ambiguity into the model due to the ignorance of context and still has room for improvement.
As for \citet{adams2017cross}, it incorporates the dictionary information into the pretraining process and then fine-tunes the model with the pretrained embeddings.
Although it also evaluates fine-tuning with the entire Long Short-Term Memory Model \cite{gers1999learning}, recent studies show Vanilla Transformer \cite{vaswani2017attention} performs better on many NMT tasks.




In this article, we propose a novel LM named Bilingual Dictionary-based Language Model (BDLM) that learns dictionary information during pretraining.
By replacing words in a sentence with information from the dictionary, BDLM could be trained on monolingual corpora.
Besides, the ambiguity problem is greatly alleviated because BDLM is only pretrained for initialization, which is fined-tuned later to better fit the NMT for word choice.
As we shall see later, the BDLM benefits the translation tasks in many aspects. 

Our main contributions are as follows:
\begin{itemize}
    \item BDLM incorporates rich information from the dictionary into the pretraining process. When being applied to NMT task, a steady increase of the BLEU score is observed.
    \item BDLM could be trained only by monolingual corpus with dictionary information, which is especially helpful for improving translation quality for low-resource languages.
    \item BDLM significantly improves the translation precision of rare words.
    \item BDLM speeds up the convergence of Vanilla Transformer for the NMT tasks.
\end{itemize}



\section{Related Work}

\paragraph{Language Model Pretraining and BERT}
\citet{devlin2018bert}'s work has shown great success in pretraining a language representation model for many natural language downstream tasks.
In this paper, two unsupervised pretraining tasks, the Masked LM (MLM) and Next Sentence Prediction (NSP), are first proposed to train the BERT.
The MLM is for getting better token representations bidirectionally while the NSP is for capturing the sentence-level relationships.
The MLM randomly masks a small ratio of tokens and set those masked tokens to be the ground truth of the corresponding sentences;
the NSP classifies the actual next sentence from a randomly selected sentence.
However, those LMs do not consider multi-lingual information for NMT.

\paragraph{Cross-lingual Language Model (XLM) Pretraining} Taking advantage of BERT's success by LM pretraining, \citet{Lample2019CrosslingualLM} proposes the idea of XLM pretraining.
They propose a Translation Language Model (TLM) to take account of multi-lingual information by simply concatenating two parallel sentences processed by MLM.
By masking words randomly in sentences of both the source side and target side, the model can attend the surrounding source context as well as the target context.
The TLM is typically used in combination with MLM.
Although it achieves several state-of-the-art results on NLU, unsupervised NMT tasks, supervised NMT tasks, and etc, it requires a large number of expensive parallel corpora, which are not available for low-resource languages.


Compared to TLM, our model is able to train on considerable monolingual corpora by introducing the dictionary information to the pretraining process.
Opposed to \citet{Ren2019ExplicitCP}, we do not explicitly build up a look-up table, which is likely to bring additional bias by the errors of the look-up table.
\citet{Ren2019ExplicitCP} proposes another XLM named cross-lingual masked language model (CMLM), which introduces a complicated N-gram embedding procedure and an N-gram translation table. 
Another similar task of ours is known as cross-lingual word embedding \cite{klementiev2012inducing}, which aims to learn cross-lingual word embeddings by pretraining with a monolingual corpus with the translation information from the dictionary.

\section{Bilingual Dictionary-based Language Model}

\begin{figure*}[ht]
\centering
\includegraphics[width=\textwidth]{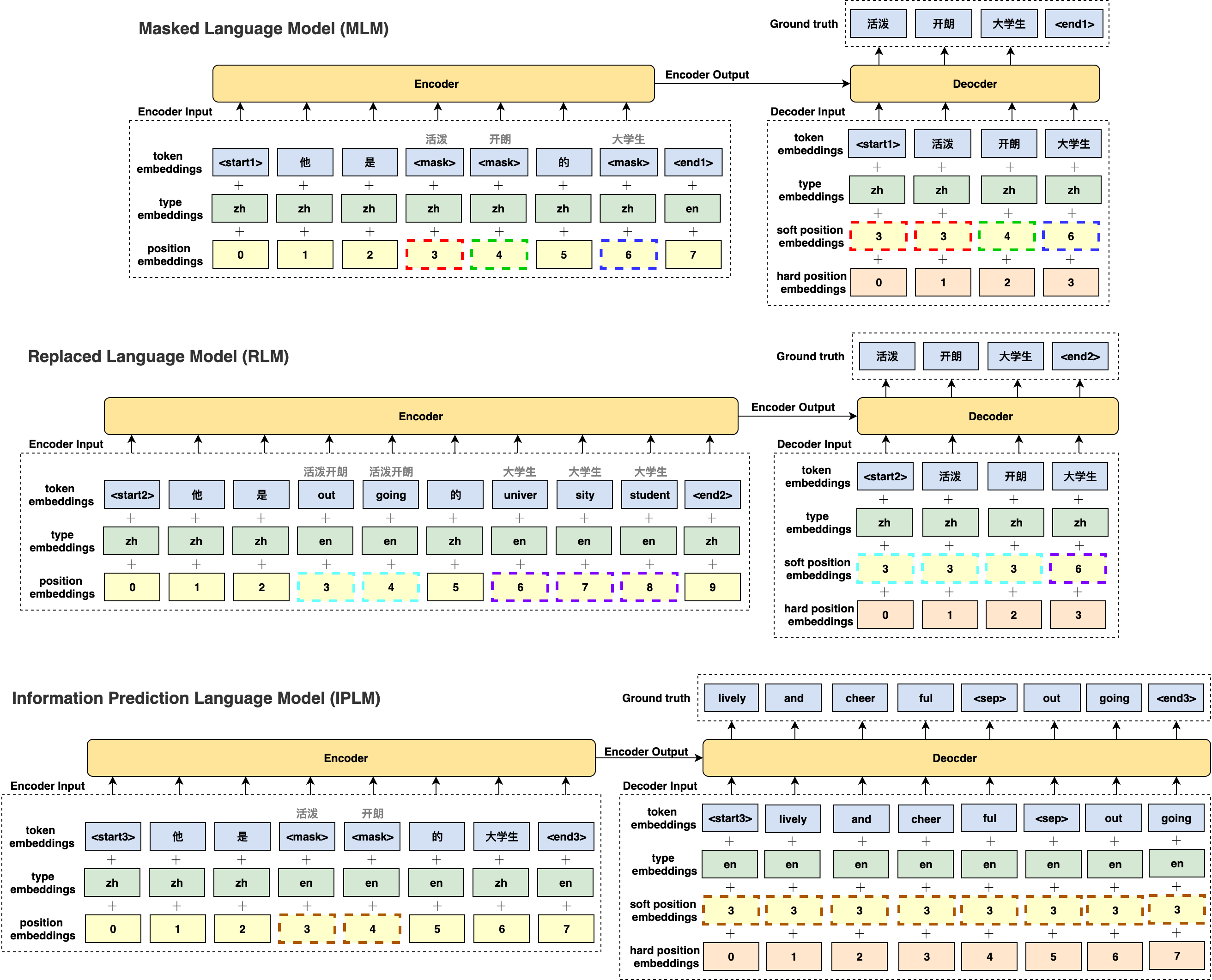}
\caption{Three LM objectives of BDLM with translation information.}
\label{fig:BDLM_translate_mode_1}
\end{figure*}


In this section, we will give a detailed introduction to the proposed BDLM.
BDLM is a multi-LM model that three LM objectives are pretrained on one unified architecture.
After pretraining, it is fine-tuned for NMT tasks with initialization with all the pretrained parameters including the embedding layers, encoder, and decoder.
No dependency on parallel corpora and representation with rich prior knowledge are its advantages.
Although BDLM can be applied to any auto-regressive encoder-decoder architecture models, we opt for implementation with Vanilla Transformer.



\subsection{Input Representations}
Following \citet{lample2017unsupervised},
we share the same vocabulary for all languages and leverage part of their input settings.
Figure \ref{fig:BDLM_translate_mode_1} demonstrates an example of the input representations during training.

\paragraph{Input of Encoder}
There are three types of embeddings at the input of the encoder: token embeddings, type embeddings, and position embeddings.
For position embeddings, we use the sinusoidal position encoding in \citet{vaswani2017attention} while the other two embeddings are learned.
The type embeddings are used to denote the information type of the corresponding token. 
It is extended from language embeddings so that BDLM can adapt to more information besides translation. 


\paragraph{Input of Decoder}

Except for token embeddings and type embeddings, there are two types of position embeddings at the input of the decoder: hard position embeddings and soft position embeddings.

The hard position embeddings denote the absolute position of each token at the decoder input sequence and are used to record the order information. 
Meanwhile, the soft position embeddings denote the first position of the masked or replaced words/phrases at the input of the encoder and are used to record their corresponding positions in the original sentence.
For example, in the IPLM inside figure \ref{fig:BDLM_translate_mode_1} (discussed in Section 3.2),
the "活泼开朗" can be translated to "lively and cheerful".
The soft positions for "lively and cheerful" are the first position of the "活泼开朗" in the encoder input, which help the model attend the "活泼开朗";
while the hard positions help specify the relative positions in the phrase "lively and cheerful".
It's worth noting that the soft position embeddings are only exploited during pre-training and will be dropped when being fine-tuned for NMT tasks.

\subsection{Multi-language-model Objectives}

BDLM contains three various LM objectives: Masked Language Model(MLM), Replaced Language Model (RLM), and Information Prediction Language Model (IPLM).
Figure \ref{fig:BDLM_translate_mode_1} shows these objectives for incorporating translation information into the pretraining process. 

\paragraph{MLM:} As proved by \citet{Lample2019CrosslingualLM}, combining MLM with other LM objectives could achieve a better representation than only using one LM objective.
We leverage the strength of multi-LM but only apply the MLM for the words that can be mapped by the dictionary.
A portion of those words is randomly masked out with special tokens [mask] and then serve as the ground truth of the corresponding input sequence. 
Then, the neural model is trained to predict the masked words from their contexts. 

\paragraph{RLM:} The difference between the MLM and the RLM is that the replaced tokens are no longer the special mask tokens.
Instead, tokens representing the information from the dictionary are placed into the masked positions.
For the case of incorporating translation information into the BDLM, we replace the masked tokens of the input sentence with the corresponding translation tokens.
In this case, the LM is forced to recover the original words from the replaced translations and the context so that the translation knowledge could be integrated into the BDLM to some extent.

\paragraph{IPLM:} Compared to the MLM, One distinct feature of the IPLM is that the LM learns to predict the linguist information of the original tokens instead of merely the original masked tokens from the context.
In case the masked token has multiple information, which is likely to happen in many scenarios like a word has multiple translations or POS.
we introduce a special separation token [sep], which aims to concatenate various information.
For the translation case of BDLM in Figure \ref{fig:BDLM_translate_mode_1}, the "活泼开朗" in the sentence of the encoder input is masked and the ground truth becomes a sequence of "lively and cheerful" and "outgoing", divided by [sep] tokens.
IPLM is trained to build up the connections between the contextual information of one word to its corresponding translation.

However, IPLM brings up ambiguity because dictionary translation does not take account of contextual information. 
The ambiguity can be alleviated because IPLM is only pretrained for initialization.
Besides, getting the knowledge of a similar representation is beneficial.
Intuitively, IPLM is a simulation of human learning a new language.
When a human learns a new language by starting with the vocabulary, he/she is probably learning non-native and imprecise knowledge.
It is easier to acquire the correct knowledge when a human gets a new language environment. 
This process maps the fine-tuning stage.
he/she can realize the vocabulary learned before is helpful in the new environment.




\subsection{Training schedule}
The multiple LM objectives are trained on one unified architecture with the same cross-entropy loss function with label smoothing \cite{szegedy2016rethinking}.
Before training, samples for each LM are randomly drawn with a fixed ratio to form a combined training set.
For the model to distinguish the various LMs, distinct special start tokens and end tokens are added to the start and the end position of the input sentences for each LM.
The dataset is then shuffled and fed into the model.
Finally, the model is able to automatically switch the pretraining process among multiple LMs according to the data being fed.



\subsection{Integrating Dictionary Information}

Figure \ref{fig:BDLM_translate_mode_1} illustrates an example of integrating translation information into BDLM.
Besides translation, BDLM can be extended to various information such as POS, synonyms, definition, and named entity (NE).
To incorporate the POS or NE information, for the RLM, we could replace the masked tokens of a word or phrase with the corresponding POS tags or NE tags.
B, M, E, O are the prefix of tags for the beginning, middle, ending, and no-chunk tokens respectively.
As for the IPLM, the ground truth becomes the sequence of those POS or NE tags.
The same methods also apply to synonyms and definitions.


\section{Experiment Settings}


\subsection{Datasets}

\begin{table}[h!]
\centering
\begin{tabular}{lrrr}
\hline
Dataset                      & Train   & Val    & Test \\ 
\hline
WMT-News'19 Zh-En            & 16k     & 2k     & 2k  \\ 
WMT'20 Zh-En$_{(NC)}$        & 300k    & 6k     & 6k  \\ 
WMT'16 Ro-En$_{(6k)}$        & 5.4k    & 0.6k   & 0.6k \\ 
WMT'16 Ro-En$_{(60k)}$       & 54k     &  6k    & 6k \\ 
\hline
\end{tabular}
\caption{\textbf{Statistics of the division of the dataset}. The WMT'20 Zh-En$_{(NC)}$ represents the WMT'20 Zh-En news-commentary dataset. Follow \citet{gu-etal-2018-universal}, the WMT'16 Ro-En$_{(6k)}$ and WMT'16 Ro-En$_{(60k)}$ are the subsets of the WMT'16 Ro-En, containing 6k and 60k sentence pairs, respectively. The "Train", "Val", and "Test" columns indicate the count of sentence pairs of the training set, validation set, and test set respectively.}
\label{tab:statistics_for_division_of_dataset}
\end{table}



The experiments were performed on WMT-News'19 Zh-En \cite{tiedemann2012parallel}, WMT'20 Zh-En news-commentary, and WMT'16 Ro-En.
The datasets contain 20K, 310K, and 610K bilingual pairs and there are 510K, 6.9M, and 14M English tokens, respectively.
Following \citet{gu-etal-2018-universal}, we only used 6k and 60k WMT'16 Ro-En sentence pairs for the training set for the purpose of low-resource testing.
Table \ref{tab:statistics_for_division_of_dataset} shows the ratio of the training, validation, and test sets of each dataset.
For the pretraining, only the training sets are used, and both the source and target sides of them are used as monolingual data.
Besides, for each set of experiments, we built a joint vocabulary with vocabulary size 50k, 80k, 90k, and 90k, respectively.
For the Chinese side, Jieba toolkit\footnote{https://github.com/fxsjy/jieba}  is used for tokenization.
After tokenization, we apply the SubwordTextEncoder toolkit in Tensorflow\footnote{https://www.tensorflow.org/datasets/api\_docs/python/\newline tfds/features/text/SubwordTextEncoder} for byte-pair encoding \cite{sennrich2015neural}.
Sentences longer than 60 subwords would be removed from the dataset.
\subsection{Dictionaries}

We integrated data from multiple sources, including Facebook research \cite{conneau2017word}, Facebook Muse\footnote{https://ai.facebook.com/tools/muse/}, Wiki-titles\footnote{http://data.statmt.org/wikititles/}, NLTK wordnet\footnote{https://www.nltk.org/howto/wordnet.html}, and ECDICT\footnote{https://github.com/skywind3000/ECDICT}.
The integrated dictionary consists of translation, Part-of-Speech (POS), synonyms, definitions, and named entity (NE) information.
We cleaned the dictionary by only preserving tokens whose translations occur in the combined corpus.
The dictionary remains 760K Chinese tokens, 1.4M English tokens, and 120K Romanian tokens after the cleaning process.
The dictionary coverage of Chinese, English and Romanian tokens are 70\%, 59\% and 27\% respectively, 
with approximately the same coverage for training, validation and test sets.


\subsection{Sample Rate}

The sample rate represents the number of repeated sampling for the same sentence during pretraining. 
For each sentence in pretraining, we perform word masking randomly. 
A higher sample rate can increase the probability that the words in each sentence are learned.

\subsection{Evaluations}
\begin{equation} \label{eq:1}
Prec_{rare/dict} = \frac{1}{N} \sum_{i=1}^{N} \frac{ \sum_{j=1}^{count(y_{i})} [y_{i,j} = \hat{y}_{i,j}]}{count(y_{i})}
\end{equation}

For the evaluation of supervised machine translation performance, we use case-insensitive BLEU score, precision for rare words (Prec$_{(rare)}$), and precision for words that appear in the dictionary (Prec$_{(dict)}$), while for the pretraining tasks, we use perplexity and accuracy. 
We used case insensitive BLEU from nltk's corpus bleu toolkit\footnote{https://www.nltk.org/\_modules/nltk/translate/bleu\_score.html}.
\citet{koehn-knowles-2017-six} serves as a good foundation for our research for metrics on rare words.
Equation \ref{eq:1} is used for Prec$_{(rare)}$ and Prec$_{(dict)}$.
The N stands for the count of words that occurs less than 10 times when calculating Prec$_{(rare)}$ and stands for the count of words appear in dictionary when calculating Prec$_{(dict)}$. 
\textbf{ y$_{i}$} and \textbf{\^y$_{i,j}$} respectively represent the ground truth and corresponding predicted words. The bracket returns 1 if \textbf{ y$_{i}$} equals to \textbf{\^y$_{i,j}$}, and returns 0 otherwise.

\subsection{Training Details}

In all experiments, following \citet{vaswani2017attention,Lample2019CrosslingualLM} we use a vanilla Transformer architecture with 8 heads, 6 layers for the encoder, 6 layers for the decoder, ReLU activations, a dropout rate of 0.1, and sinusoidal positional embeddings. 
For the reasons of training time and the NMT performance of Vanilla Transformer on WMT-News'19 Zh-En, we choose 128 of hidden units and of embedding size.
The training set for NMT is used for pretraining.
We train our models with the Adam optimizer \cite{kingma2014adam}, set the initial learning rate to 1e-4 and the mini-batch size is 16. 
To build a competitive baseline model, we use a shared token embedding layer and type embedding layer described in section 3 for both encoder and decoder.
Besides, Label smoothing with $\epsilon=0.1$ is added for the cross-entropy loss.
Greedy decoding is used and the average accuracy of the translated tokens in the validation set is used as a stopping criterion for training.


\section{Results}

In this section, we empirically present BDLM's fine-tuning results and demonstrate the effectiveness of NMT, which shows our approach significantly outperforms Vanilla Transformer and TLM.

\subsection{Golden information in Dictionary}

\begin{table}[ht]
\centering
\begin{tabular}{lccc}
\hline
models               & BLEU   & Prec$_{(rare)}$ & Prec$_{(dict)}$ \\ \hline
Vanilla              & 46.6     & 46.5    & 54.2   \\ 
\hline
MLM                  & 46.7     & 46.8    & 54.3   \\ 
TLM                  & 48.8     &  47.5   & 55.8   \\ 
\hline
translate            & \textbf{55.0} & \textbf{53.7} & \textbf{61.4} \\
POS                  &   44.5   &  45.5    & 52.3      \\ 
synonyms             &   47.7   &  48.0    & 56.0      \\
definitions          &   37.4   &  34.4    & 42.2      \\ 
NE                  & 51.0     &   52.2   & 59.2      \\
\hline
\end{tabular}
\caption{\textbf{Supervised NMT Results on WMT-News'19 Zh-En.} BLEU, precision of rare words, and precision of words in the dictionary are used for evaluation.
Models pretrained with different schemes are evaluated.
Vanilla Transformers without pretraining, with MLM pretraining, and with TLM pretraining are used as baselines.
For BDLM, we investigate translation, POS, synonym, definition, and NE information to examine which information is beneficial to NMT.}
\label{tab:result_on_wmt_news}
\end{table}

\begin{table}[ht]
\centering
\begin{tabular}{lcccc}
\hline
tasks & size & epochs & acc & ppl \\ 
\hline
MLM      &   80k  &  5  &  5.2  &  3.08     \\ 
TLM      & 91k  &   11   &  14.9  &  5.14  \\ 
\hline
translate &  132k  &  80  & 13.3 &  2.06  \\
POS       & 71k &  8  & 12.1 &  2.28   \\ 
synonyms  &  12k & 66 & 7.4    &  2.15         \\
definitions & 23k  &  \textbf{165}  & \textbf{48.8} &  2.25   \\ 
NE       &  \textbf{142k}  &   48  &  17.2  &  \textbf{2.03}      \\
\hline
\end{tabular}
\caption{\textbf{Statistics and Performance of Different pretrain Tasks.} For each model, we included the training data size, number of epochs the pretraining runs, pretraining's token-level accuracy, and perplexity.}
\label{tab:pretrain_wmt_news}
\end{table}

A typical bilingual dictionary contains rich information includes but not limited to: translation, POS, synonyms, definition, and named entity (NE). 
The translation information naturally fits the NMT task because the pretraining process could build a connection between the translation representation and the corresponding context;
synonyms could also lead to a better-generalized model with diversified translations; NE information helps the model to locate and recognize the entity.

In Table \ref{tab:result_on_wmt_news}, we evaluate five BDLMs, each one leverages one distinct information from the dictionary, which is the translation, POS, synonym, definition, and NE information, respectively.
To prove the benefit of incorporating information from dictionary into pretraining, 
we include three baselines: Vanilla Transformers without pretrain, with MLM pretrained, and with TLM pretrained.
We evaluate on WMT-News'19 Zh-En and use BLEU, precision of rare words, and precision of words in the dictionary as metrics.
Our results show that BDLM with translation information (BDLM$_{translate}$) gets a superior performance of 55.0 BLEU over other BDLMs, which proves that the translation information is the most valuable for NMT.
Besides BDLM$_{translate}$, BDLM$_{NE}$ also significantly outperforms the three baselines.
To better interpret the results, Table \ref{tab:pretrain_wmt_news} shows the statistics and performance of the models.
The data sizes vary among different tasks due to the various size of information; the epochs vary because the models early stop at different epochs automatically.
One of the factors that contribute to the better performance of BDLM$_{translate}$ and BDLM$_{NE}$ is their large data size.
However, although BDLM$_{synonym}$ has the smallest data size, it still improves the Vanilla and MLM by 1.0+ BLEU; BDLM$_{translate}$ has the second-largest data size but performs the best on NMT.
It proves that the data size is related to the effect of pretraining, but the relevance to the translation task itself is more important.
The BDLM$_{POS}$ performs poorly because of the under-fitting problem.
Due to the noisiness and small size of data, only 8 epochs were trained for BDLM$_{POS}$.
Meanwhile, the BDLM$_{definition}$ suffers from the over-fitting problem.
BDLM$_{definition}$ trained too many epochs and the discrepancy between the definition and translation information is huge, leading to worse performance than Vanilla.
Additionally, our empirical experience shows the token-level accuracy and perplexity of pretraining have no direct relationship with the NMT performance but only demonstrate how well the task has been trained.

\subsection{Effect of Sample Rate}

\begin{table}[ht]
\centering
\begin{tabular}{lll}
\hline
& \multicolumn{2}{l}{$sample rate$} \\
$models$      & 3.0   & 10.0 \\ \hline
BDLM$_{translation}$ & 38.4 & \textbf{55.0} \\
BDLM$_{NE}$         & 41.2 & \textbf{51.0} \\ 
\hline
\end{tabular}
\caption{\textbf{BLEU for BDLMs with Different Sample Rate.} We tested BDLM$_{translate}$ and BDLM$_{NE}$ with a sample rate of 3.0 and 10.0.}
\label{tab:sample_rate}
\end{table}

In Table \ref{tab:sample_rate}, we highlight experiments on translation and NE models with a sample rate of 3.0 and 10.0.
Based on experimental results, a higher sample rate yields better performance. 

\subsection{Evaluation for Convergence Speed}

    

\begin{figure}[h!]
    
        \centering
        \includegraphics[width=\linewidth]{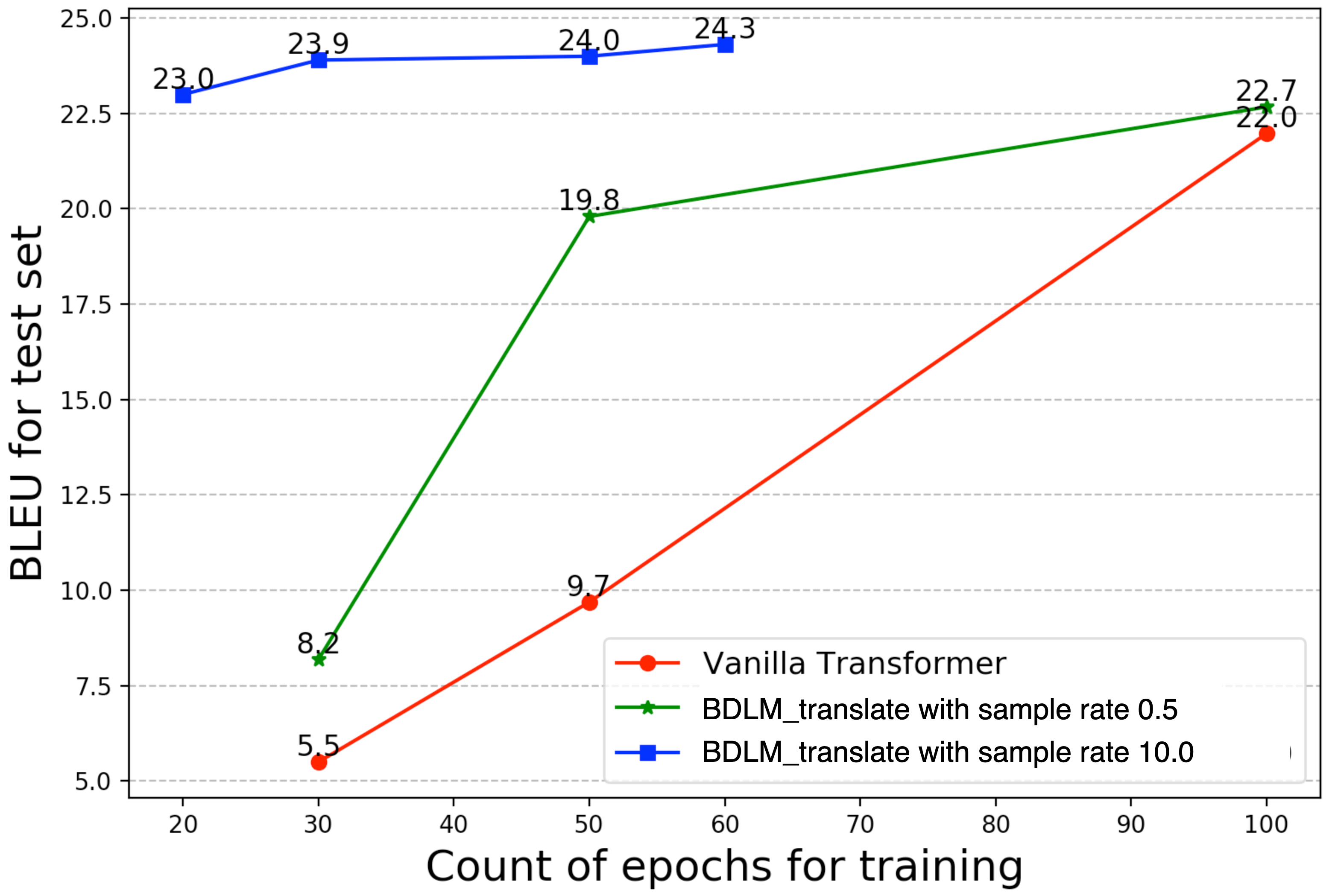}
        \caption{\textbf{The Convergence Speed Comparison on WMT’20 Zh-En news-commentary}. We Calculate the BLEU at multiple epochs during training.}
        \label{fig:convergence_speed}
\end{figure}

To measure the effectiveness of pretraining, we compared the Convergence Speed of BDLM$_{translate}$ with Vanilla. 
The train and test data are from WMT’20 Zh-En news-commentary.
Figure \ref{fig:convergence_speed} shows the obvious advantages of BDLM$_{translate}$ in convergence speed. 
BDLM$_{translate-10.0}$ achieves 23.0 BLEU at the 20th epoch, which is higher than BDLM$_{translate-0.5}$ and Vanilla at the 100th epoch. 
It implies that the model has learned abundant effective representation during pretraining.



   

\subsection{Effect on Translating Rare words}

\begin{figure*}[ht]
\centering
\includegraphics[width=\textwidth]{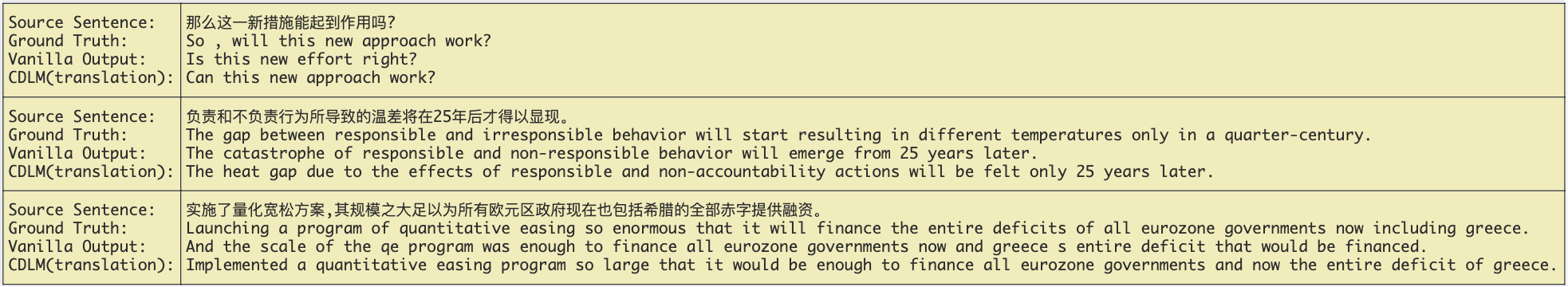}
\caption{\textbf{Examples of Translation Results.} We extracted 3 examples from test data to show how the BDLM$_{translate-10.0}$ helps the translation quality.  }
\label{fig:BDLM_example}
\end{figure*}

\begin{figure}[h!]
    \centering{
    {\includegraphics[width=\linewidth]{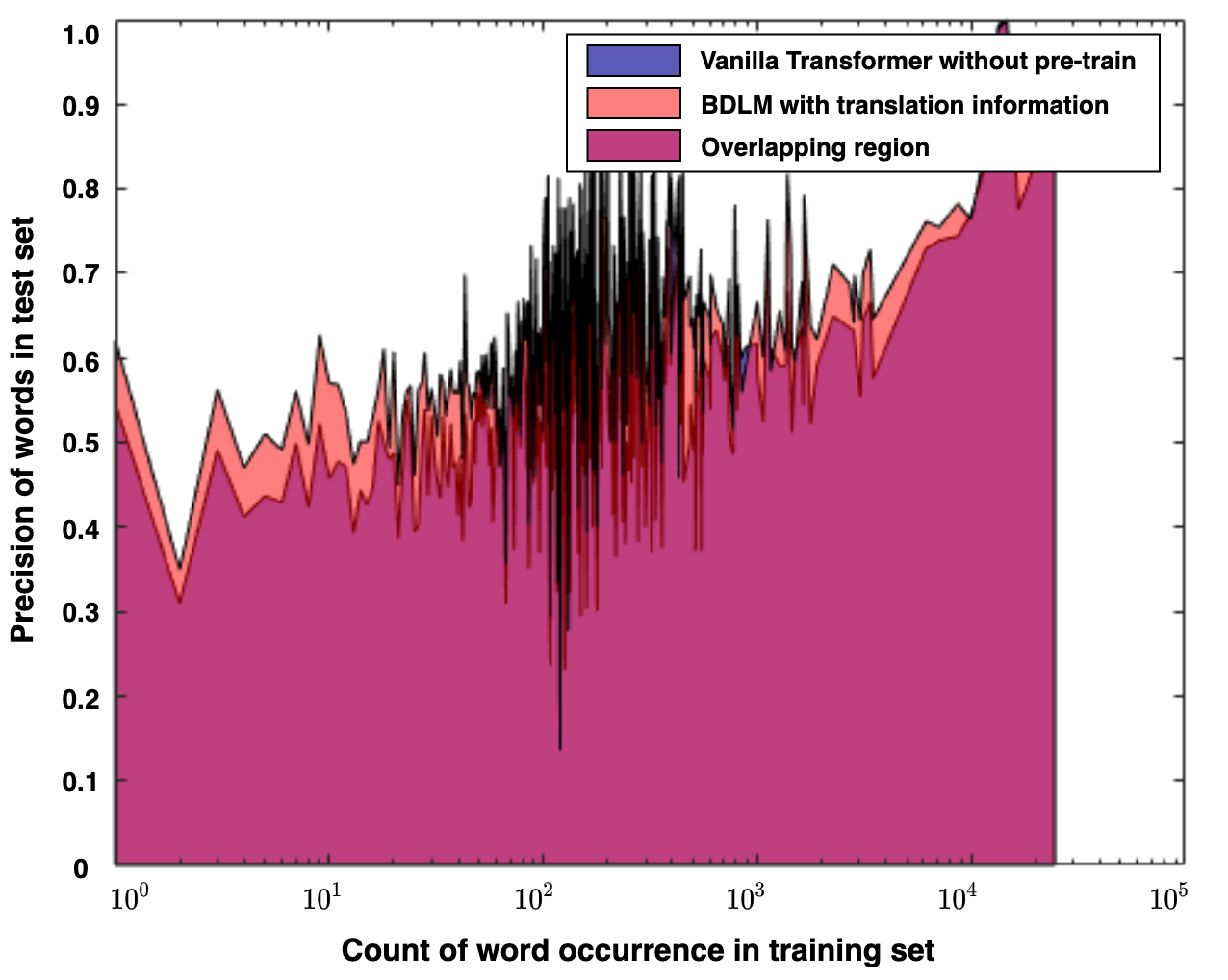}}
    \caption{\textbf{Precision of the translated words.} The left portion of the graph corresponds to rarer words. The blue indicates Vanilla Transformer model performed better, the red indicates the BDLM$_{translate}$ performed better, and the magenta is their overlapping regions.}
    \label{fig:precision}
    }
\end{figure}



We further evaluate the precision of translated words with different occurrence frequencies with the BDLM$_{translate}$ in section 5.1.
As shown in Figure \ref{fig:precision}, it compares the precision performance of the Vanilla Transformer with the BDLM$_{translate}$.
The x-axis of the figure represents the training dataset occurrence groups and is set at a logarithmic scale. 
The left portion of the graph corresponds to rarer words. 
The y-axis corresponds to the ratio of words found in the predicted sentences compared to the ground truth. 
The red region is always on the top of the magenta region and there is a bigger red area on the left portion of the figure, which confirms that BDLM$_{translate}$ is effective for improving the translation of rare words.
In addition, the precision of rare words and precision of words in the dictionary shown in Table \ref{tab:result_on_wmt_news} also proves that BDLM$_{translate}$ and BDLM$_{NE}$ have large gains on the translation of rare words.

Through the model trained in section 5.3, we use some test cases to observe the advantages of BDLM in Rare words translation. 
Figure~\ref{fig:BDLM_example} shows translation sentence samples from BDLM$_{translation-10.0}$ and Vanilla.



\begin{figure*}[h!]
    \centering
    \includegraphics[width=\linewidth]{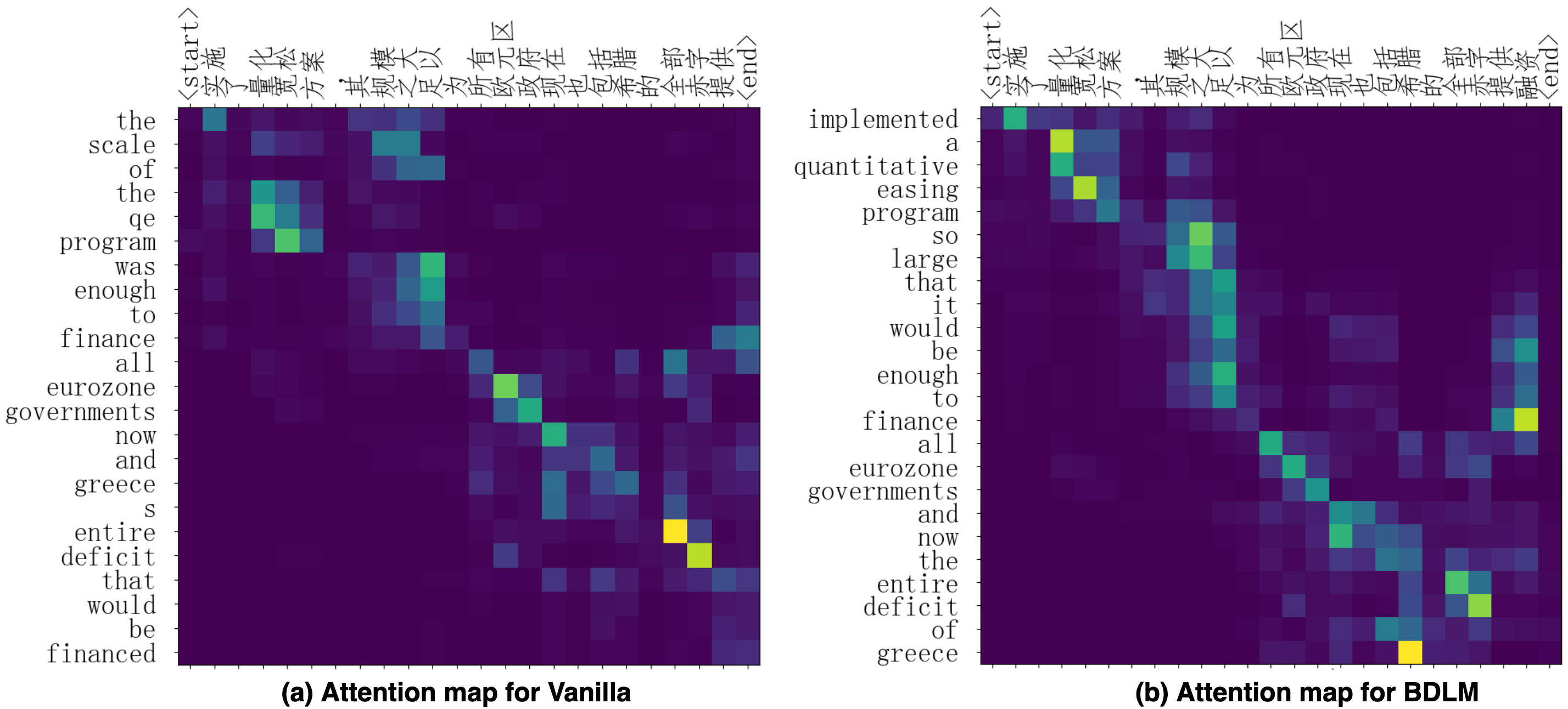}
    \caption{Attention map for the 3rd example in Figure \ref{fig:BDLM_example}}
    \label{fig:attention_map}
\end{figure*}




In the third example, the Chinese phrase "实施了量化宽松方案" can be successfully translated to "implemented a quantitative easing program" in BDLM$_{translate-10.0}$ while Vanilla cannot. 
Four words in the phrase "实施", "量化", "宽松" and "方案" all appear in the dictionary used in pretraining. 
This indicates that pretraining improves the accuracy of rare words translation.

Figures \ref{fig:attention_map}  show the attention maps for a Zh-En translation example.
The lighter the region is, the larger the attention value is. Figure \ref{fig:attention_map}.a uses Vanilla for translation while Figure \ref{fig:attention_map}.b uses BDLM$_{translate-10.0}$ in section 5.3.
When comparing the two figures, we can find that the diagonal brightness of Figure \ref{fig:attention_map}.b is higher than that of Figure \ref{fig:attention_map}.a. 
Especially the upper left corner area, where the "量化宽松方案" locates, BDLM$_{translate-10.0}$'s attention is more observable.

\subsection{Effect for Low-resource Language}

\begin{table*}[ht!]
    \centering
    \begin{tabular}{ccccccc}
    \hline
    models & ro-en$_{6k}$ & ro-en$_{60k}$ & en-ro$_{6k}$ & en-ro$_{60k}$ \\
    \hline 
        
        Vanilla & 3.6 & 33.3 & 3.5 & 30.7 \\
        BDLM & 5.4 & \textbf{35.6} & 5.2 & 32.9 \\
        Vanilla\cite{gu-etal-2018-universal}  & 1.2 & 12.1 & -- & -- \\
        Multi-NMT+UnivTok\cite{gu-etal-2018-universal} & \textbf{20.1} & 24.3 & -- & -- \\
    \hline
    \end{tabular}
    \caption{BLEU for NMT on WMT'16 Ro-En}
    \label{tab:my_label}
\end{table*}

In previous experiments, we only verified the effectiveness of BDLM on  
zh and en. 
Both Chinese and English are high-resource languages, meaning that we can obtain a large number of parallel corpus and dictionary resources. 
\citet{gu-etal-2018-universal} uses Multi-lingual NMT with a transfer-learning method to share lexical and sentence level representations across multiple source languages into one target language.
The main goal is to help low-resource languages by sharing the learned model. 
We adapted their experimental settings to measure the performance of BDLM in a low-resource language.
We use WMT'16 Ro-En data as a monolingual resource to pretrain the model, and train NMT models on 6k and 60k sentence pairs respectively. 
In addition, we tested our Vanilla and BDLM on the En-Ro task to verify the generalization of the model.
As shown in Table \ref{tab:my_label}, due to our baseline boost settings described in Section 4.4, our Vanilla exceeds \citet{gu-etal-2018-universal}'s Vanilla significantly.
The BDLM outperforms Vanilla in every experiment. 
In the 6k training setting, Multi-NMT+UnivTok's performance is phenomenal because they used a large number of parallel corpora of other high-resource languages. However, the BDLM's advantages are shown in the 60k training setting. It outperforms Multi-NMT+UnivTok by 11.3 BLEU.
BDLM and Multi-NMT+UnivTok both significantly improve the performance of low-resource language. 

\section{Conclusion}

We proposed a novel BDLM that incorporates dictionary information into the pretraining process.
We investigated multiple information, including translation, POS, synonyms, definitions, and NE information, and prove that the translation information is the most valuable for NMT tasks.
Compared with TLM, our proposed BDLM does not need any parallel corpus.
Our experiments show that BDLM pretraining is effective for improving the BLEU scores for NMT tasks.
Due to random masking during pretraining, a higher sample rate can increase the probability that each word of the sentence is learned, thereby improving the BLEU value of the model.
In addition, BDLM's convergence speed is significant faster than Vanilla Transformer. 
When predicting rare words, BDLM also has obvious improvement for words that appear in the dictionary.
Through experiments of WMT'16 Ro-En, we also verified the considerable improvement of BDLM in a low-resource language NMT task.


\section{Future Work}

We are interested in conducting ablation experiments for the three pretrain LMs of BDLM to understand how these LMs benefit NMT independently.
Besides, we plan to use a large amount of data for pretraining with the support of more GPUs.
We also plan to apply a multi-task for incorporating the translation, NE, and synonym information into the pretraining process.
Finally, extending the bilingual to multi-lingual model, 
especially for those low-resource languages, is in the schedule.

\bibliography{anthology,eacl2021}

\begin{thebibliography}{16}
\expandafter\ifx\csname natexlab\endcsname\relax\def\natexlab#1{#1}\fi

\bibitem[{Adams et~al.(2017)Adams, Makarucha, Neubig, Bird, and
  Cohn}]{adams2017cross}
Oliver Adams, Adam Makarucha, Graham Neubig, Steven Bird, and Trevor Cohn.
  2017.
\newblock Cross-lingual word embeddings for low-resource language modeling.
\newblock In \emph{Proceedings of the 15th Conference of the European Chapter
  of the Association for Computational Linguistics: Volume 1, Long Papers},
  pages 937--947.

\bibitem[{Conneau et~al.(2017)Conneau, Lample, Ranzato, Denoyer, and
  Jégou}]{conneau2017word}
Alexis Conneau, Guillaume Lample, Marc'Aurelio Ranzato, Ludovic Denoyer, and
  Hervé Jégou. 2017.
\newblock \href {http://arxiv.org/abs/1710.04087} {Word translation without
  parallel data}.

\bibitem[{Devlin et~al.(2018)Devlin, Chang, Lee, and
  Toutanova}]{devlin2018bert}
Jacob Devlin, Ming-Wei Chang, Kenton Lee, and Kristina Toutanova. 2018.
\newblock Bert: Pre-training of deep bidirectional transformers for language
  understanding.
\newblock \emph{arXiv preprint arXiv:1810.04805}.

\bibitem[{Gers et~al.(1999)Gers, Schmidhuber, and Cummins}]{gers1999learning}
Felix~A Gers, J{\"u}rgen Schmidhuber, and Fred Cummins. 1999.
\newblock Learning to forget: Continual prediction with lstm.

\bibitem[{Gu et~al.(2018)Gu, Hassan, Devlin, and Li}]{gu-etal-2018-universal}
Jiatao Gu, Hany Hassan, Jacob Devlin, and Victor~O.K. Li. 2018.
\newblock \href {https://doi.org/10.18653/v1/N18-1032} {Universal neural
  machine translation for extremely low resource languages}.
\newblock In \emph{Proceedings of the 2018 Conference of the North {A}merican
  Chapter of the Association for Computational Linguistics: Human Language
  Technologies, Volume 1 (Long Papers)}, pages 344--354, New Orleans,
  Louisiana. Association for Computational Linguistics.

\bibitem[{Kingma and Ba(2014)}]{kingma2014adam}
Diederik~P Kingma and Jimmy Ba. 2014.
\newblock Adam: A method for stochastic optimization.
\newblock \emph{arXiv preprint arXiv:1412.6980}.

\bibitem[{Klementiev et~al.(2012)Klementiev, Titov, and
  Bhattarai}]{klementiev2012inducing}
Alexandre Klementiev, Ivan Titov, and Binod Bhattarai. 2012.
\newblock Inducing crosslingual distributed representations of words.
\newblock In \emph{Proceedings of COLING 2012}, pages 1459--1474.

\bibitem[{Koehn and Knowles(2017)}]{koehn-knowles-2017-six}
Philipp Koehn and Rebecca Knowles. 2017.
\newblock \href {https://doi.org/10.18653/v1/W17-3204} {Six challenges for
  neural machine translation}.
\newblock In \emph{Proceedings of the First Workshop on Neural Machine
  Translation}, pages 28--39, Vancouver. Association for Computational
  Linguistics.

\bibitem[{Lample and Conneau(2019)}]{Lample2019CrosslingualLM}
Guillaume Lample and Alexis Conneau. 2019.
\newblock Cross-lingual language model pretraining.
\newblock In \emph{NeurIPS}.

\bibitem[{Lample et~al.(2017)Lample, Conneau, Denoyer, and
  Ranzato}]{lample2017unsupervised}
Guillaume Lample, Alexis Conneau, Ludovic Denoyer, and Marc'Aurelio Ranzato.
  2017.
\newblock Unsupervised machine translation using monolingual corpora only.
\newblock \emph{arXiv preprint arXiv:1711.00043}.

\bibitem[{Ren et~al.(2019)Ren, Wu, Liu, Zhou, and Ma}]{Ren2019ExplicitCP}
Shuo Ren, Yu~Wu, Shujie Liu, Ming Zhou, and Shuai Ma. 2019.
\newblock Explicit cross-lingual pre-training for unsupervised machine
  translation.
\newblock In \emph{EMNLP/IJCNLP}.

\bibitem[{Sennrich et~al.(2015)Sennrich, Haddow, and
  Birch}]{sennrich2015neural}
Rico Sennrich, Barry Haddow, and Alexandra Birch. 2015.
\newblock Neural machine translation of rare words with subword units.
\newblock \emph{arXiv preprint arXiv:1508.07909}.

\bibitem[{Szegedy et~al.(2016)Szegedy, Vanhoucke, Ioffe, Shlens, and
  Wojna}]{szegedy2016rethinking}
Christian Szegedy, Vincent Vanhoucke, Sergey Ioffe, Jon Shlens, and Zbigniew
  Wojna. 2016.
\newblock Rethinking the inception architecture for computer vision.
\newblock In \emph{Proceedings of the IEEE conference on computer vision and
  pattern recognition}, pages 2818--2826.

\bibitem[{Tiedemann(2012)}]{tiedemann2012parallel}
J{\"o}rg Tiedemann. 2012.
\newblock Parallel data, tools and interfaces in opus.
\newblock In \emph{Lrec}, volume 2012, pages 2214--2218.

\bibitem[{Vaswani et~al.(2017)Vaswani, Shazeer, Parmar, Uszkoreit, Jones,
  Gomez, Kaiser, and Polosukhin}]{vaswani2017attention}
Ashish Vaswani, Noam Shazeer, Niki Parmar, Jakob Uszkoreit, Llion Jones,
  Aidan~N Gomez, {\L}ukasz Kaiser, and Illia Polosukhin. 2017.
\newblock Attention is all you need.
\newblock In \emph{Advances in neural information processing systems}, pages
  5998--6008.

\bibitem[{Zhang and Zong(2016)}]{Zhang2016}
Jiajun Zhang and Chengqing Zong. 2016.
\newblock Bridging neural machine translation and bilingual dictionaries.
\newblock \emph{ArXiv}, abs/1610.07272.

\end{thebibliography}
\bibliographystyle{acl_natbib}

\end{CJK*}
\end{document}